\documentclass[twocolumn,letterpaper,10pt]{article}
\usepackage{kuz_EN}
\usepackage{graphicx}				
\usepackage[utf8]{inputenc}  				
\usepackage[round,authoryear]{natbib}
\usepackage[english]{babel}					
\usepackage[labelfont=bf]{caption}			
\usepackage[colorlinks=false]{hyperref} 	

\makeatletter
\def\convertto#1#2{\strip@pt\dimexpr #2*65536/\number\dimexpr 1#1}
\makeatother

\begin{document}

\date{}
\title{Safety of human-robot interaction through tactile sensors and peripersonal space representations}

\author{
\vspace{1ex}
\textbf{Petr Švarný, Matěj Hoffmann} \\ 
Czech Technical University in Prague, Faculty of Electrical Engineering, Department of Cybernetics\\
Karlovo náměstí 13, 121 35 Prague 2\\
Email: petr.svarny@fel.cvut.cz, matej.hoffmann@fel.cvut.cz \\
}

\maketitle 

\thispagestyle{empty}

\defcitealias{ISO12100}{ISO~12100}
\defcitealias{ISO10218}{ISO~10218}
\defcitealias{ISO/TS15066}{ISO/TS~15066}
\defcitealias{ISO13849}{ISO~13849}

\noindent
{ \begin{center} \bf\normalsize Abstract\end{center}}
{
\noindent
Human-robot collaboration including close physical human-robot interaction (pHRI) is a current trend in industry and also science. The safety guidelines prescribe two modes of safety: (i) power and force limitation and (ii) speed and separation monitoring. We examine the potential of robots equipped with artificial sensitive skin and a protective safety zone around it (peripersonal space) to safe pHRI.
} 

\section{Introduction}
The combination of words safety and robotics often incites images of a machine uprising in the minds of laymen. 
Contemporary robotics, however, faces a great deal of challenges connected to even mundane interaction scenarios between robots and humans. 

\section{Standardization}
The overall safety in physical human-robot interaction (pHRI) is subject to many standards. These start with the general machinery standards like \citetalias{ISO12100}, \citetalias{ISO13849}, followed by specific robot standards as \citetalias{ISO10218}. The speed of robotics evolution makes standardization very difficult. The newest standard in preparation \citetalias{ISO/TS15066} mirrors the trend of collaborative robotics, but still has a lot of discussion ahead before it can become an accepted standard~\cite{haddadin2015physical}\cite{jacobs2018flourishing}.

\section{Collaborative operation}
Robots have to adhere to all the mechanical safety standards as any other industrial machinery. However, as opposed to classical machines, robots can have complex behavior while interacting with people. 

The \citetalias{ISO10218} and \citetalias{ISO/TS15066} specify four types of safe pHRI:
\begin{enumerate}
	\item Safety-rated monitored stop
    \item Hand guiding
    \item Power and force limiting by inherent design or control
    \item Speed and separation monitoring
\end{enumerate}

In the first two regimes, there is no simultaneous autonomous movement of the robot and the human collaborator allowed: in 1), the robot will stop whenever the human enters the workspace; in 2), the robot operates in a specific hand-guiding (kinesthetic teaching) mode and does not execute any independent movements. The other two regimes, on the other hand, constitute the real challenge.

\begin{figure}
	\centering
	\includegraphics[width=2.5in]{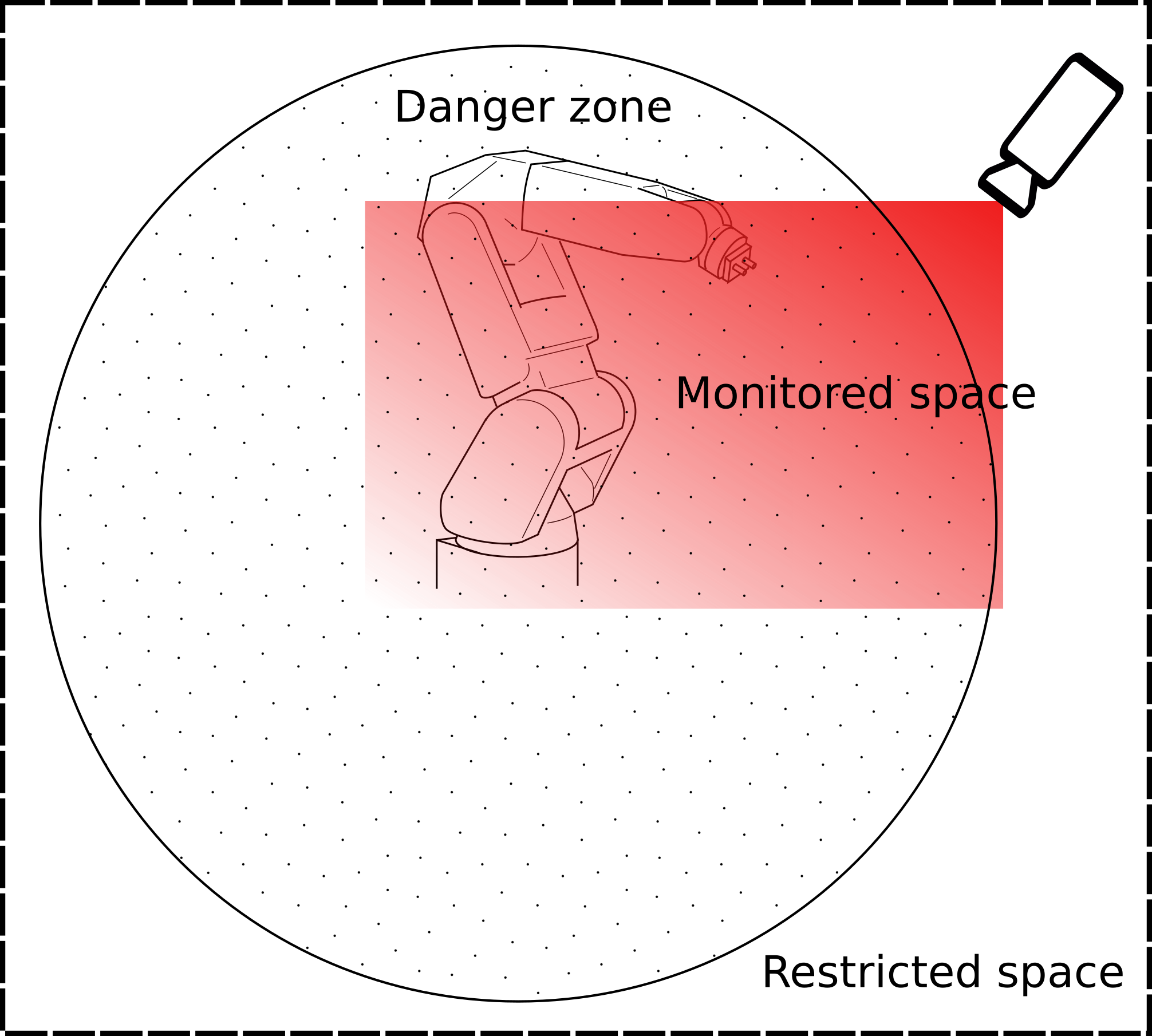}
	\caption{Robot in a monitored space.}
	\label{fig:robot_space}
\end{figure}

\subsection{Power and force limiting}
Power and force limiting allows physical contacts with a moving robot but they need to be within human body part specific limits on force, pressure, and energy. Example of the safety foundation on the robot side is a lightweight structure. The perception of collisions leads to appropriate reactions (e.g., \citep{magrini2015control}). A recent survey on this \textit{post-impact} interaction control is \citep{haddadin2017robot}. 

\subsection{Speed and separation monitoring}
Speed and separation monitoring deals with \textit{pre-impact} interaction. It demands reliable estimation of distances between robots and humans. Proper estimation allows the alteration of the robots behavior in order to maintain the minimal separation distance between the operator and the robot that cannot be crossed. However, light curtains or safety-rated  scanners (e.g., SafetyEYE\footnote{http://pilz.com}) are very coarse (monitor 2D or 3D zones) whereas sensor with higher resolution (e.g., cameras or RGB-D sensors) from which also human skeleton can be extracted are currently not safety-rated \cite{flacco2015depth,Nguyen2018}.

\begin{figure}
	\centering
	\includegraphics[width=2.5in]{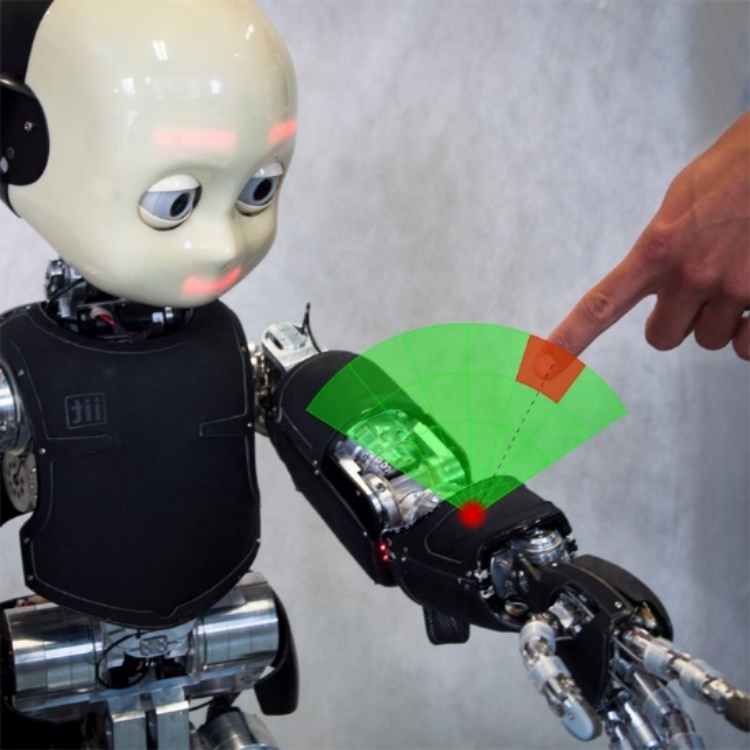}
	\caption{Schematics of a receptive field that is part of the peripersonal space of the iCub robot~\citep{roncone2016peripersonal}.}
	\label{fig:iCub}
\end{figure}

\section{Artificial skins and peripersonal space}
Our own research uses robots with pressure-sensitive skins, like the iCub. These can be exploited for contact detection and response but also for calibration of the safety margin through visuo-tactile associations (see Fig. \ref{fig:iCub}) and \citep{roncone2016peripersonal}). Alternatively, the safety margin can rely on distal sensing using cameras or RGB-D sensors and human skeleton extraction by convolutional neural networks. The availability of safety-rated human keypoint extraction or at least point cloud detection would dramatically expand the possibilities of human-robot collaboration in the speed and separation monitoring regime.

\section{Conclusion}
Safe pHRI is a dynamically evolving field with some borders set by industry standards but with a vivid discussion about best practices.

\vspace{5ex}
\noindent
{\bf\large Acknowledgement}\vspace{2ex} \newline
{Petr Švarný was supported by the Grant Agency of the Czech Technical University in Prague, grant No. SGS18/138/OHK3/2T/13. Matěj Hoffmann was supported by the Czech Science Foundation under Project GA17-15697Y.}

\vspace{3,16314mm}
\nocite{*}
\bibliographystyle{apalike_kuz_EN}
\bibliography{references}

\end{document}